\documentclass[10pt,twocolumn,letterpaper]{article}

\usepackage[pagenumbers]{cvpr} 

\usepackage{xcolor}
\usepackage{colortbl}
\usepackage{amssymb} 
\usepackage{pifont}  
\usepackage{dsfont}
\usepackage{algorithm}        
\usepackage{algpseudocode}    

\newcommand{\tick}{\textcolor{green}{\checkmark}}
\newcommand{\xmark}{\textcolor{red}{\ding{55}}}

\definecolor{bestgreen}{RGB}{120,200,120}
\definecolor{secondgreen}{RGB}{180,230,180}
\definecolor{thirdyellow}{RGB}{255,255,180}

\newcommand{\best}[1]{\cellcolor{bestgreen}\textbf{#1}}
\newcommand{\second}[1]{\cellcolor{secondgreen}#1}
\newcommand{\third}[1]{\cellcolor{thirdyellow}#1}

\DeclareMathOperator*{\argmin}{arg\,min}
\DeclareMathOperator*{\argmax}{arg\,max}

\definecolor{cvprblue}{rgb}{0.21,0.49,0.74}
\usepackage[pagebackref,breaklinks,colorlinks,allcolors=cvprblue]{hyperref}
\usepackage{acronym}

\def\paperID{33389} 
\def\confName{CVPR}
\def\confYear{2026}

\title{ ACE-SLAM: Scene Coordinate Regression for Neural Implicit Real-Time SLAM}

\author{Ignacio Alzugaray \quad Marwan Taher \quad Andrew J. Davison\\
 {\normalsize Dyson Robotics Lab,  Imperial College London}\\
{\tt\small \{i.alzugaray,m.taher,a.davison\}@imperial.ac.uk}
}
\begin{document}

\maketitle


\begin{abstract}
We present a novel neural RGB-D \ac{SLAM} system that learns an implicit map of the scene in real time. For the first time, we explore the use of \ac{SCR} as the core implicit map representation in a neural \ac{SLAM} pipeline, a paradigm that trains a lightweight network to directly map 2D image features to 3D global coordinates. 
\ac{SCR} networks provide efficient, low-memory 3D map representations, enable extremely fast relocalization, and inherently preserve privacy, making them particularly suitable for neural implicit SLAM.

Our system is the first one to achieve strict real-time in neural implicit RGB-D \ac{SLAM} by relying on a \ac{SCR}-based representation. 
We introduce a novel \ac{SCR} architecture specifically tailored for this purpose and detail the critical design choices required to integrate \ac{SCR} into a live SLAM pipeline. 
The resulting framework is simple yet flexible, seamlessly supporting both sparse and dense features, and operates reliably in dynamic environments without special adaptation.
We evaluate our approach on established synthetic and real-world benchmarks, demonstrating competitive performance against the state of the art.
\end{abstract}
    
\acrodef{SCR}{Scene Coordinate Regression}
\acrodef{ACE}{Accelerated Scene Coordinate Regression}
\acrodef{SLAM}{Simultaneous Localization And Mapping}
\acrodef{NeRF}{Neural Radiance Fields}

{\small \textbf{Project Page: \href{https://ialzugaray.github.io/ace-slam/}{https://ialzugaray.github.io/ace-slam/}}}

\section{Introduction}
\label{sec:intro}
The choice of scene representation is fundamental in any 3D Spatial AI algorithm, as this models the constraints between geometry, appearance, and motion.
Each scene representation has characteristics and limitations which are inherently propagated through the pipeline.
In visual \ac{SLAM}, efficiency and robustness are often just as important as accuracy, because embodied edge applications such as autonomous navigation, manipulation or wearables usually impose hard real-time constraints.

\begin{figure}
    \centering
\includegraphics[width=0.7\linewidth]{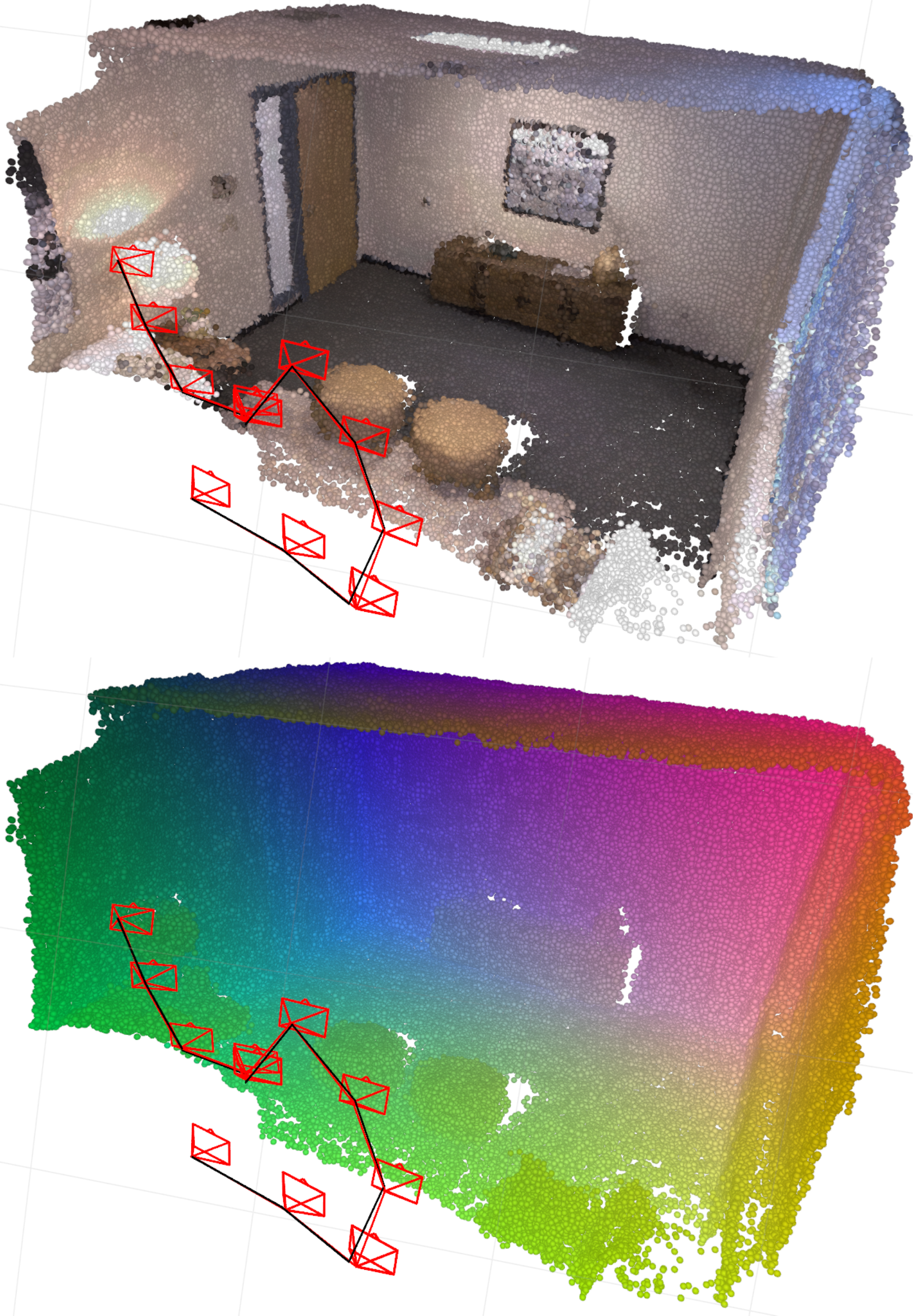}
\caption{Our system takes RGB-D frames, extracts 2D image features, and directly regresses their 3D global coordinates using a novel \ac{SCR} architecture that implicitly models the scene. 
This highly-compressive implicit map (here 1MB) can be optimized at test time in our proposed neural \ac{SLAM} system, ACE-SLAM, to simultaneously estimate geometry (top) in global coordinates (bottom) and optimize camera trajectory (red, ground truth in black) in real time. The system provides native relocalization capabilities and is robust to dynamic elements in the scene.}
    \label{fig:teaser}
\end{figure}

Scene representations in Visual \ac{SLAM} have evolved rapidly, adopting new ideas from Computer 
Vision and testing their trade-offs in terms of efficiency and expressiveness.
Early visual \ac{SLAM} pipelines relied on explicit geometric primitives that could be effectively handled using the limited compute capacity of CPUs such as points~\cite{Engel2018DSO,Campos2021ORB_SLAM3,MurArtal2017ORB_SLAM2,Qin2018VINSMono}, lines~\cite{GomezOjeda2019PLSLAM,Pumarola2017PLSLAM}, or volumetric grids~\cite{oleynikova2017voxblox,Grinvald2019VoxbloxPlusPlus}.
The advent and commoditization of GPUs enabled hybrid pipelines with neural components. These components have often been used to provide inputs like depth, optical flow or pose estimation~\cite{Tateno2017CNN-SLAM,Teed2021DROIDSLAM,Murai2025MASt3RSLAM} to be used with standard scene representations. It was also found that GPU-accelerated primitives could be used themselves as new types of explicit map representation such as Gaussian splats and related methods~\cite{Kerbl2023GaussianSplatting,Held2025Triangle,Matsuki2024GaussianSplattingSLAM}. 

More intriguing still  have been visual \ac{SLAM} methods which use neural networks directly as scene representations. These neural implicit techniques represent the detailed geometry of a scene 
as neural codes or tokens ~\cite{Bloesch2018CodeSLAM,Czarnowski2020DeepFactors} or straightforwardly as the weights of a neural network~\cite{sucar2021imap,Zhu2022NICE-SLAM,Zhu2024NICER} which is incrementally optimised at run-time.
Many of these latter approaches have been designed around volumetric rendering pipelines following the rise of \ac{NeRF} \cite{Mildenhall2020NeRF, Muller2022InstantNGP}. 
\ac{NeRF} was originally designed for novel view synthesis, while in \ac{SLAM}, efficient and accurate localization and scene geometry is usually the focus.
Extracting accurate geometry from  implicit volumetric representations requires significant computational effort as integration via sampling along camera rays is required, hampering the real-time performance of many implicit neural \ac{SLAM} systems.


Building on these observations, in this paper we investigate the use of Scene Coordinate Regression (SCR)~\cite{shotton2013scr_forest} as an alternative implicit neural scene representation well-suited to the real-time and localization-oriented demands of \ac{SLAM}.
In \ac{SCR}, a small neural network is trained to encode information about a scene, but in a way which is quite different from rendering-based approaches like \ac{NeRF}. 
A \ac{SCR} network is trained to directly regresses global 3D coordinates from the input of 2D image features~\cite{brachmann2023ace,brachmann2024acezero}, bypassing the need for ray integration and enabling instantaneous relocalization. 
While \ac{SCR} has proven robust and efficient for large-scale offline mapping~\cite{Wang2024glace,jiang2025rscore}, with resilience to outliers and inherent geometric consistency, its potential as a unified representation for continuous, real-time mapping and tracking from live streams remains largely unexplored.

We leverage \ac{SCR} as the core mapping and localization component of ACE-SLAM, a neural RGB-D system that constructs compact (a few MBs), scene-specific implicit maps online via self-supervision and continual learning (see \cref{fig:teaser}).
By performing mapping and tracking directly through \ac{SCR}, ACE-SLAM achieves accurate, and real-time operation on consumer-grade hardware, with relocalization and implicit loop closure emerging naturally from the formulation.
The learned scene representation is also privacy-preserving by design, as the stored geometric information can only be accessed through corresponding images captured within the scene.
These characteristics enable a simple, unified neural implicit \ac{SLAM} pipeline without the need for additional subsystems, demonstrating competitive performance and establishing a new paradigm for real-time implicit visual SLAM. 
In summary, the contributions of this paper are:
\begin{itemize}
\item The first \textbf{RGB-D SLAM pipeline} that fully leverage a \ac{SCR}-based implicit map representation for \textbf{strict real-time} operation, capable of processing live data at frame rate, unlike previous neural implicit SLAM methods.
\item A \textbf{novel SCR network architecture} specifically designed for continuous online training and inference within the SLAM loop.
\item A detailed analysis of design choices and efficiency trade-offs when adopting SCR representations for practical, real-time SLAM.
\item Extensive evaluation on established SLAM benchmarks, including \textbf{dynamic scenes}, demonstrating the resilience and competitive performance of the proposed system against other \textbf{neural implicit RGB-D SLAM pipelines}.
\end{itemize}

\section{Related Work}
\acrodef{RANSAC}{Random Sample Consensus}
\acrodef{DSAC}{Differentiable Sample Consensus}
\acrodef{ACE}{Accelerated Coordinate Encoder}
\textbf{The Rise of Neural SLAM.}
Early \ac{SLAM} systems relied on explicit geometric primitives such as points~\cite{Engel2018DSO,Campos2021ORB_SLAM3,MurArtal2017ORB_SLAM2,Qin2018VINSMono}, lines~\cite{GomezOjeda2019PLSLAM,Pumarola2017PLSLAM}, and volumetric grids~\cite{oleynikova2017voxblox,Grinvald2019VoxbloxPlusPlus} to represent scenes and  motion. 
In recent years, neural priors have been incorporated to improve tasks such as depth, optical flow, and pose estimation~\cite{Tateno2017CNN-SLAM,Teed2021DROIDSLAM,Murai2025MASt3RSLAM,Lipson2024DPVSLAM}, or even provide single-shot inference for both motion and geometry~\cite{Wang2025VGGT, Wang2025Pi3, Liu2025SLAM3R, Wang2025CUT3R}. 
While these are generally accurate and robust, they are also computationally expensive approaches that heavily rely on large pre-trained priors.
At the same time, explicit primitive-based representations have resurged with Gaussian- or Triangle-splatting\cite{Kerbl2023GaussianSplatting,Held2025Triangle,Matsuki2024GaussianSplattingSLAM}, providing more efficient online performance with minimal priors, but yielding generally larger maps and limited representation capacity depending on the primitive of choice.

\textbf{Implicit Neural RGB-D SLAM.}
Early attempts to integrate optimizable implicit neural scene representations into \ac{SLAM} pipelines \cite{Czarnowski2020DeepFactors,Bloesch2018CodeSLAM} saw limited adoption. 
\ac{NeRF} \cite{Mildenhall2020NeRF} transformed the field by enabling neural implicit maps to be optimized via self-supervision  with inverse rendering, and this technique was first adopted for visual SLAM by iMAP \cite{sucar2021imap}.
Variations on this initial method for neural implicit RGB-D \ac{SLAM} included using voxel-based feature grids \cite{Muller2022InstantNGP} in NICE-SLAM~\cite{Zhu2022NICE-SLAM}, feature planes~\cite{Chen2022TensoRF} in ESLAM~\cite{Johari2023eslam}, or Point-\ac{NeRF}~\cite{Xu2022PointNeRF} in Point-SLAM~\cite{Sandstrom2023pointslam}.
These methods tightly couple tracking and mapping through inverse rendering, often leading to slow convergence which makes online \ac{SLAM} challenging \cite{Tosi2024ReshapingSLAM}.
Most efforts to address this have decoupled the tracking stage, so that local camera motion and/or 2D correspondences are first resolved and optimized before mapping \cite{Zhang2023goslam,Wang2024StructerfSLAM,Yin2024IBDSLAM,Deng2024neslam}. 
Additional components are typically needed for loop closure via global or local bundle adjustment and pose graph optimization strategies \cite{Wang2023coslam,Zhang2023goslam}. 
Handling dynamic scenes usually requires specialized modules that segment moving regions, often using semantics, to avoid corrupting the implicit map \cite{Xu2024nidslam,Jiang2024rodynslam,Wang2025fndslam,Li2025ddnslam}.
In contrast, by adopting a simpler \ac{SCR}-based representation, our approach achieves real-time operation, native loop closure, and robust handling of dynamic regions within a unified, streamlined SLAM formulation.

\textbf{Scene Coordinate Regression.}
\ac{SCR} was first introduced in \cite{shotton2013scr_forest} as a learned map using a random forest regression model, enabling fast camera relocalization via \ac{RANSAC}~\cite{Fischler1981RANSAC}. Early neural network–based \ac{SCR} approaches emerged alongside the \ac{DSAC} works \cite{Brachmann2019NeuralGuidedRANSAC,Brachmann2019ExpertSampleConsensus,brachmann2021dsacstar}, but these methods were slow to converge, often requiring hours for high-quality maps.
A major breakthrough came with improved training strategies leveraging sampling and self-supervised data augmentations \cite{Wang2024HSCNetPlusPlus}, culminating in \ac{ACE} \cite{brachmann2023ace}, which reduced training times to minutes.
Since then, \ac{SCR} techniques have gained traction, particularly for modelling large-scale scenes \cite{Wang2024glace,jiang2025rscore,liu2025mace}, incorporating geometric priors \cite{Bian2025SceneCoordinateReconstructionPriors}, enhancing relocalization \cite{Chai2015OptimizedACE}, and improving generalization beyond scene-specific training \cite{Bruns2025ACEG}. These methods typically rely on ground-truth poses to first map the scene offline before enabling efficient online relocalization.
More closely related to our work, \ac{ACE}-Zero \cite{brachmann2024acezero} operates without known poses, alternating incrementally between mapping and relocalization to produce a \ac{SCR} model from unordered frames. 
Inspired by this, we propose ACE-SLAM, which adapts \ac{SCR} to continuous RGB-D camera streams, requiring careful design to ensure robustness and stability while performing simultaneous mapping and tracking in real-time.

\section{Methodology}
\acrodef{MLP}{Multilayer Perceptron}

\newcommand{\keypoint}{{\mathbf x}} 
\newcommand{\image}{{\mathcal{I}}}
\newcommand{\depthmap}{\mathcal{D}}
\newcommand{\map}{{\mathcal{M}}}
\newcommand{\intrinsics}{{\mathbf{K}}}
\newcommand{\encoder}{{\mathcal{F}}}
\newcommand{\feature}{{\mathbf f}}
\newcommand{\keypoints}{{\mathbf X}} 
\newcommand{\point}{{\mathbf y}} 
\newcommand{\R}{{\mathbb R}} 
\newcommand{\depth}{d} 
\newcommand{\globalcoord}{\tilde{\point}} 
\newcommand{\localcoord}{\point} 
\newcommand{\project}{\boldsymbol{\pi}} 
\newcommand{\unproject}{\boldsymbol{\pi}^{-1}} 
\newcommand{\SEthree}{\textbf{SE}(3)}
\newcommand{\pose}{\mathbf{P}}
\newcommand{\inlierratiothresh}{\tau}
\newcommand{\curly}[1]{\left\{#1\right\}}
\newcommand{\PoseSolver}{\mathcal{G}}
\newcommand{\InlierRatio}{\mathbf{\lambda}}
\newcommand{\InlierRatioFn}{\InlierRatio}
\newcommand{\Inlier}{\mathds{1}}
\newcommand{\NumHypothesis}{H}
\newcommand{\NumFrame}{M}
\newcommand{\NumFeatures}{M}
\newcommand{\residual}{r}
\newcommand{\ResidualFun}{r}
\newcommand{\parameters}{\Theta}
\newcommand{\poses}{\mathcal{P}}
\newcommand{\loss}{\mathcal{L}_\localcoord}
\newcommand{\trimlp}{TriMLP}
\newcommand{\acemlp}{HomMLP}

\newcommand{\InsertionTimeThresh}{\bar{\sigma}}
\newcommand{\InsertionInlierThresh}{\bar{\InlierRatio}}
\newcommand{\BatchSize}{B}
\newcommand{\NumBatches}{N_B}

\newcommand{\OptimizationWindow
}{\mathcal{W}}
\newcommand{\NewestKeyframeWindow
}{W_L}
\newcommand{\OldKeyframeWindow
}{W_G}

\begin{figure}
    \centering
    \includegraphics[width=0.8\linewidth]{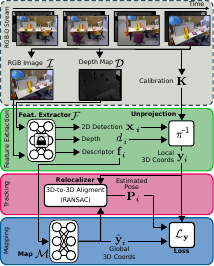}
    \caption{ACE-SLAM system overview described in \cref{sec:system}}
    \label{fig:pipeline}
\end{figure}

\subsection{Scene Coordinate 
Regression Formulation}
\label{sec:formulation}
\textbf{Preliminaries.}
The input to our system (see \cref{fig:pipeline}) is a stream of RGB-D pairs 
$\{\image^t, \depthmap^t\}$, where color $\image^t \in \mathbb{R}^{H\times W\times 3}$ 
and depth $\depthmap^t \in \mathbb{R}^{H\times W}$ images are captured at time $t$ with known camera intrinsics $\intrinsics$.
Each RGB-D pair is processed by a feature extractor $\encoder$ (see \cref{sec:system}),
producing a set of $\NumFeatures$ visual features:
\begin{align}
\encoder^t = \encoder(\image^t, \depthmap^t) = \{\encoder^t_i\} = 
\{(\feature^t_i, \keypoint^t_i, \depth^t_i)\}_{i=1}^{\NumFeatures}~,
\end{align}
where $\keypoint^t_i \in \mathbb{R}^2$ denotes the 2D keypoint location,
$\depth^t_i \in \mathbb{R}$ its corresponding depth from $\depthmap^t$,
and $\feature^t_i \in \mathbb{R}^D$ the local feature descriptor.

Each keypoint is unprojected into its 3D \emph{local} coordinate using the camera intrinsics as  
$\localcoord_i^t = \unproject_\intrinsics(\keypoint_i^t, \depth_i^t)$,  
yielding the set of feature–geometry pairs  
$\mathcal{F}^t = \{(\feature_i^t, \localcoord_i^t)\}_{i=1}^{\NumFeatures}$,  
which represent image features and their corresponding 3D \emph{local} coordinates for subsequent SCR-based mapping and localization.

\textbf{Implicit Scene Representation from \ac{SCR}.}
The implicit map $\map$ is a scene-specific trainable neural network (see \cref{sec:triplane}) that regresses 3D \emph{global} coordinates $\globalcoord_i = \map(\feature_i) \in \R^3$ directly from 2D image descriptors $\feature_i$. 
This operation is fully parallelizable per feature and independent across pixels, a property that is key for efficient inference. 
When applied to all extracted features $\curly{\feature_i^t}$ from a frame $\curly{\image^t, \depthmap^t}$, the network independently predicts the corresponding global scene coordinates $\curly{\globalcoord_i^t}$. 
Such direct regression establishes a one-to-one mapping between image features and 3D global points, eliminating the need for the ray-based volumetric sampling required in rendering-based neural implicit \ac{SLAM} formulations.

\textbf{SLAM formulation.}
Our goal in this paper is to perform \ac{SLAM} by jointly optimizing a set of camera poses $\poses = \{\pose^0, \ldots, \pose^\NumFrame\} \in \SEthree$ and the parameters of the neural implicit map $\map$. 
Each residual $\residual_i^t$ measures the geometric consistency between the predicted 3D global scene coordinate $\globalcoord_i^t = \map(\feature_i^t)$ and its corresponding 3D local observation $\localcoord_i^t = \unproject_\intrinsics(\keypoint_i^t, \depth_i^t)$ transformed by the estimated pose $\pose^t$. 
The objective is to minimize the sum of these residuals across all frames and features, enforcing mutual consistency between map and poses:
\begin{align}
\residual_i^t(\map,\pose^t) &= \|\map(\feature_i^t) - \pose^t \localcoord_i^t\|^2, \label{eq:residual}\\
\loss(\map,\poses) &= \sum_{\pose^t \in \poses}\sum_{\encoder_i^t \in \encoder^t} \residual_i^t(\map,\pose^t)
,\\
\{\map^*, \poses^*\} &= \argmin_{\map, \poses} \loss(\map,\poses).
\end{align}
Here, the loss $\loss$ aggregates per-frame losses composed of pixel-wise residuals within each image.
This formulation is inherently self-supervised, enforcing geometric consistency between the learned implicit map and the estimated poses using only RGB-D observations, with no external ground-truth supervision. 
Notably, there is no explicit notion of an ``image'', only collections of independent pixel-level constraints coupled through shared poses --- enabling flexible sampling strategies to accommodate real-time constraints.

\subsection{\trimlp{}: SCR by Triplane Coordinate Voting}
\label{sec:triplane}

Conventional \ac{SCR} networks typically employ simple \acp{MLP} that directly regress 3D coordinates from input features.  ACE
\cite{brachmann2023ace,brachmann2024acezero} employs a architecture with fully-connected layers with skip connections and prediction of homogenous coordinates, which we refer to as \acemlp{} and compare against in \cref{sec:experiments}.
Other have opted for additional refinement modules \cite{Wang2024glace,jiang2025rscore}. 
Here, we propose a lightweight triplane-based alternative, referred to as \trimlp, that factorizes 3D coordinate regression into three orthogonal 2D classification problems.

Given a feature descriptor $\feature_i$, a compact \ac{MLP} first predicts classification logits over discretized spatial bases on the three orthogonal planes $(XY, XZ, YZ)$:
\begin{align}
C_i^{XY}, C_i^{XZ}, C_i^{YZ} = \mathrm{softmax}(\mathrm{MLP}(\feature_i)),
\end{align}
where $C_i^{IJ} \in \R^{r_I \times r_J}$ represents a normalized voting distribution over the grid of basis coordinates $B^{IJ} \in \R^{r_I \times r_J \times 2}$ on the corresponding plane.

Each plane independently contributes coordinate estimates via weighted averaging:
\begin{align}
(\tilde{x}_i^{XZ}, \tilde{z}_i^{XZ}) &= \sum B^{XZ} \odot C_i^{XZ}, \\
(\tilde{x}_i^{XY}, \tilde{y}_i^{XY}) &= \sum B^{XY} \odot C_i^{XY}, \\
(\tilde{y}_i^{YZ}, \tilde{z}_i^{YZ}) &= \sum B^{YZ} \odot C_i^{YZ},
\end{align}
and the final 3D global coordinate is computed by averaging compatible components across planes:
\begin{align}
\globalcoord_i &= 
\begin{bmatrix}
\tilde{x}_i \\
\tilde{y}_i \\
\tilde{z}_i
\end{bmatrix} 
=
\frac{1}{2}
\begin{bmatrix}
\tilde{x}_i^{XY} + \tilde{x}_i^{XZ} \\
\tilde{y}_i^{XY} + \tilde{y}_i^{YZ} \\
\tilde{z}_i^{XZ} + \tilde{z}_i^{YZ}
\end{bmatrix} = \mathrm{\trimlp}(\feature_i).
\end{align}

\begin{figure}
    \centering
    \includegraphics[width=\linewidth]{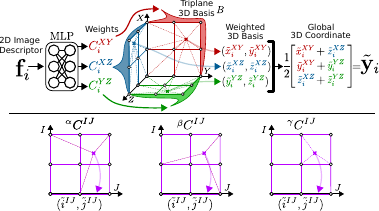}
    \caption{
The proposed \trimlp{} \ac{SCR} performs coordinate regression by voting over a set of predefined 3D bases $B$ across three orthogonal planes (Top).
The same bases $B^{IJ}$ can be combined with different voting weights (${}^{\alpha}C^{IJ}$, ${}^{\beta}C^{IJ}$, ${}^{\gamma}C^{IJ}$) while still producing the same voting location $\left(\tilde{i}^{IJ}, \tilde{j}^{IJ}\right)$ (Bottom). 
This explicit many-to-one mapping facilitates fast adaptation of the network to new incoming data.
}
    \label{fig:triplane}
\end{figure}

This formulation preserves spatial locality and interpretability while embedding an inductive bias that accelerates convergence and improves test-time adaptability, which is critical for online real-time \ac{SLAM}. 
This triplane\footnote{Unlike triplanes used  in neural rendering for feature interpolation and decoding, our \trimlp{} directly votes for 3D coordinates via  classification.}
 design allows features that correspond to the same 3D location but have differing descriptors (\eg, a corner viewed from multiple angles) to find multiple valid paths to the same global coordinate (see \cref{fig:triplane}). 
This flexibility is particularly important for shallow, narrow MLPs, which can otherwise excessively entangle features and hinder online adaptation.
We compare the performance of the \trimlp{} and conventional \acemlp{} in \cref{sec:experiments}.

\subsection{ACE-SLAM}
\label{sec:system}
Frames from the RGB-D input stream are immediately processed upon acquisition by a  frozen feature extractor (see \cref{sec:formulation}) in a separate thread and then fed to our system.

Our system runs continuously using a parallel tracking-and-mapping  strategy~\cite{Klein2007PTAM}, alternating between between the estimation of the poses $\poses$ and the neural implicit map $\map$ in optimization cycles, only interrupted by the asynchronous insertion of new frames fed by the feature extractor thread. 
The steps of these cycles are described below.

\textbf{Pose Estimation via Relocalization.}
Our pipeline fully relies on relocalization to estimate camera poses. 
Let the features be $\mathcal{F}^t = \{ (\feature_i^t, \localcoord_i^t) \}_{i=1}^{\NumFeatures}$ extracted from frame $t$, and their corresponding global coordinates (assuming $\map$ is fixed) be $\globalcoord_i^t = \map(\feature_i^t)$. The  camera pose $\pose^t$ is estimated as the minimization of the  rigid alignment error:
\begin{align}
\pose^t &= \argmin_{\pose \in \SEthree} \sum_{i=1}^N \residual_i^t(\map, \pose)
= \argmin_{\pose \in \SEthree} \sum_{i=1}^N \|\globalcoord_i^t - \pose \localcoord_i^t\|^2, \label{eq:pose_refinement}
\end{align}
which can be solved efficiently in closed form using the Kabsch-Umeyama algorithm~\cite{Umeyama1991LeastSquares}. To cope with outliers in the global coordinate estimates $\globalcoord_i^t$, we apply \ac{RANSAC}~\cite{Fischler1981RANSAC}.  
Triplets of correspondences are sampled to propose up to $\NumHypothesis$ candidate poses hypotheses $\poses^\mathcal{H}\in \SEthree$, and the one with the largest inlier ratio $\InlierRatio^t$ is selected:
\begin{align}
\InlierRatio^t(\pose) &= \frac{1}{N} \sum_{i=1}^N \mathbf{1}\big(\residual_i^t(\map, \pose) < \inlierratiothresh\big),\\
\pose^t &= \argmax_{\pose \in \poses^\mathcal{H}} \InlierRatioFn(\pose)~,
\end{align}
with a pre-specified inlier ratio threshold $\inlierratiothresh$.
The best hypothesis is further refined using only inliers $\left(\curly{(\globalcoord^t_i,\localcoord^t_i) \,|\, \residual_i^t(\map,\pose^t) < \inlierratiothresh}\right)$ with \cref{eq:pose_refinement}. 
This relocalization procedure  not only provides the best-fit pose among the hypotheses, but also quantifies the quality of the estimation as the inlier ratio $\InlierRatio^t$. 
Low values indicate either tracking failure or map degradation (\eg, due to forgetting).
This is leveraged to dynamically allocate resources to frames requiring more correction as detailed later.

This always-on relocalization strategy is markedly more efficient than differential rendering-based tracking, which requires volumetric integration, as is used in rendering-based neural implicit \ac{SLAM} systems. 
Unlike gradient-based methods, it also produces estimates without the need for a good pose prior, enabling frame skipping for real-time operation while inherently supporting loop closure and robustness to scene dynamics (see \cref{sec:experiments}).

\textbf{Keyframing Strategy and Optimization Window.}  
The keyframes are a subset of all incoming frames, 
whose poses $\poses$ are actively refined and whose features are used to train the map $\map$.
Any incoming frame $t$ is first relocalized against the up-to-date map $\map$, producing an initial pose estimate $\pose^t$ and inlier ratio $\InlierRatio^t$. 
A frame becomes a keyframe if more than  $\InsertionTimeThresh$ time has elapsed since the last keyframe insertion, ensuring scene coverage, or if its inlier ratio falls below a threshold $\InlierRatio^t < \InsertionInlierThresh$, signalling the risk of losing track (\eg if we explore faster than the map is able to adapt).

Inspired by \cite{sucar2021imap}, we select a subset of keyframes in each optimization cycle to perform tracking and mapping, denoted as the optimization  window $\OptimizationWindow$. 
This always includes the newest  $\NewestKeyframeWindow$ keyframes to enforce accurate local estimation.
Moreover, the latest frame, regardless whether is promoted to a keyframe, is also included, to facilitate the continual tracking allowing rapid adaptation to newly observed scene regions. 
Beyond these, up to $\OldKeyframeWindow$ other keyframes sampled.
Each frame $t$ is added to $\OptimizationWindow$ with a probability $p^t \propto \tfrac{1}{|\poses|} + \alpha (1 - \InlierRatio^t)$, where the first term encourages uniform coverage and the second favours frames with low inlier ratios, \ie those that were poorly aligned when last relocalized, with $\alpha$ controlling the exploration-exploitation balance (here $\alpha=1$). 
The inclusion of these other keyframes in $\OptimizationWindow$ implicitly lead to soft loop closure capabilities and drive the global consistency of the estimated solution. 

In each optimization cycle, an optimization window $\OptimizationWindow$ is sampled. 
The frames in $\OptimizationWindow$ are first relocalized (and their inlier ratios $\InlierRatio^t$ updated) and then used for mapping, as detailed next. 

\textbf{Feature Sampling and Neural Implicit Mapping.} 
For each frame in $\OptimizationWindow$, a subset of their features is sampled for map optimization. 
Similarly to the selection strategy for $\OptimizationWindow$, the probability $p_i^t$ of sampling a specific feature $\feature_i^t$ is inversely proportional to the relocalization inlier ratio of its originating frame $t$, combined with a uniform component, 
$p_i^t \propto \tfrac{1}{\sum_i |\encoder^i|} + \beta(1 - \InlierRatio^t)$, with $\beta=1$. 
This strategy samples more features from the most uncertain frames, dynamically allocating compute where the map requires greater refinement.

At each optimization iteration, a set of sampled features with descriptors $\feature_i^j$, local coordinates $\localcoord_i^j$, and associated poses $\pose_i^j$ (previously relocalized and assumed fixed at this stage) forms the optimization set $\mathcal{S}_\OptimizationWindow = \curly{(\feature_i^j, \localcoord_i^j, \pose_i^j)}$. 
The map is then refined by minimizing the residuals:
\begin{align}
\map^* = \argmin_{\map} \sum_{(\feature_i^j, \localcoord_i^j, \pose_i^j) \in \mathcal{S}_\OptimizationWindow} \|\map(\feature_i^j) - \pose_i^j \localcoord_i^j\|^2. 
\end{align}
In practice, we use mini-batches of size $\BatchSize$ and perform stocastic gradient descent over a few $\NumBatches$ batches per optimization cycle. 
This rarely exhausts the available samples within $\OptimizationWindow$ (\ie completes in fractions of a second), but quickly iterating in fresh optimization cycles allows updated map information to promptly propagate to the relocalization stage, enabling the system to rapidly adapt to scene changes.

Throughout the pipeline, no explicit feature correspondences or frame-to-frame constraints are used and, instead, frames interact implicitly through the implicit neural map representation. 
A compact, representation-constrained map (\eg, a small \ac{MLP}) promotes clustering of similar features in 3D, leading to emergent implicit matching and information transfer across distant regions of the map. 
This naturally yields soft loop-closure behavior without additional modules, progressively enforcing global consistency -- an aspect that could be made enforced explicit in future work.

Note that the compute cost remains fixed throughout all stages of the pipeline: a constant number of frames are sampled in the optimization window $\OptimizationWindow$, each relocalized using a fixed number of $\NumHypothesis$ pose hypotheses, and a fixed set of samples $\mathcal{S}_\OptimizationWindow$ is processed per cycle. 
This design ensures predictable iteration cost and stable real-time performance.

\textbf{Feature Extraction.}
While our framework is agnostic to the choice of feature extractor, it desired that features are descriptive, repeatable, have good spatial coverage and, most importantly, can be extracted efficiently.
The pipeline does not explicit match features, and thus it naturally supports both dense or sparse features, allowing the system to operate with  different performance–accuracy trade-offs.

In this work, we employ the \textit{ACE} feature encoder~\cite{brachmann2023ace, brachmann2021dsacstar}, which produces dense feature maps at $1/8$ resolution. 
Alternatively, we also evaluate our pipeline using \textit{SuperPoint} (SP)~\cite{detone2018superpoint}, which produces a sparse set of high-quality features.
All feature extractors are kept frozen (pre-trained) throughout our experiments, while their online test-time refinement remain an interesting direction for future work.
Using these feature extractors, only the color image from the RGB-D input is used, while depth information is ignored. 
Future extensions could incorporate depth to produce true RGB-D features or even leverage depth priors, as in ACEZero~\cite{brachmann2024acezero}, enabling operation on RGB-only streams.

\acrodef{ATE}{Absolute Trajectory Error}
\acrodef{RMSE}{Root mean Square Error}

\section{Experimental Evaluation}
\label{sec:experiments}

We evaluate our proposed neural implicit RGB-D \ac{SLAM} system against state-of-the-art methods, highlighting the advantages of adopting \ac{SCR} as an implicit map representation. 
This formulation offers a strong alternative to neural rendering–based approaches, demonstrating efficient, flexible, and compact mapping capabilities.

All experiments were conducted on an NVIDIA RTX 4090 GPU, 64\,GB RAM and an Intel i7-12700K CPU. 
Reported metrics are averaged across 3 runs to account  for stochasticity.
Additional implementation details, parameter settings, and results are provided in the supp. material.

\textbf{Datasets.} 
The \textit{Replica} dataset~\cite{Straub2019Replica} provides high-quality synthetic indoor scenes with accurate ground-truth geometry and trajectories, from which we adopt the sequences used in~\cite{sucar2021imap}. 
For real-world evaluation, we use the \textit{TUM-RGBD}~\cite{Sturm2012BenchmarkRGBD} and \textit{ScanNet}~\cite{Dai2017ScanNet} datasets and the sequences commonly evaluated in the literature~\cite{Tosi2024ReshapingSLAM}.

\textbf{Metrics for Trajectory and Geometry.} 
We evaluate trajectory accuracy using \ac{ATE} \ac{RMSE} against ground-truth trajectories with EVO~\cite{grupp2017evo}. 
As our system targets strict real-time operation, evaluation is performed on the trajectory estimated just after the last frame is received, without any additional refinement. 
The estimated trajectory corresponds to all keyframes relocalized against the final map at completion time. 
For fairness, no post-processing is applied and all  keyframes, including those with low inlier ratios are retained, although they could be straightforwardly detected.

Scene geometry shown in figures correspond to the inferred 3D global coordinates of inlier features from all keyframes, as detected by the system. 
We emphasize that our quantitative evaluation focuses on trajectory accuracy rather than high-fidelity geometry, aiming for maps of sufficient quality to enable real-time \ac{SLAM}, rather than neural rendering for novel-view synthesis as pursued by other approaches in the literature. 
Nonetheless, additional results on geometry are provided in the supplementary material.

\begin{table*}[t]
\centering
\scriptsize
\setlength{\tabcolsep}{2pt}
\renewcommand{\arraystretch}{1.2}
\caption{\textbf{\ac{ATE} \ac{RMSE} [m] ($\downarrow$) in static Replica, TUM-RGBD, and ScanNet datasets}. Methods using decoupled tracking Front-End (FE), Bundle Adjustment (BA), or explicit Loop Clousure (LC) are noted. Entries are replicated from \cite{Tosi2024ReshapingSLAM}. X indicates failure to complete.}
\begin{tabular}{l|ccc|cccccccc|ccc|cccccc}
\hline
\textbf{Method} & \textbf{FE} & \textbf{BA} & \textbf{LC} &
\multicolumn{8}{c|}{\textbf{Replica}} & 
\multicolumn{3}{c|}{\textbf{TUM}} & 
\multicolumn{6}{c}{\textbf{ScanNet}} \\
\cline{2-21}
 & & & &
R0 & R1 & R2 & O0 & O1 & O2 & O3 & O4 
& fr1/desk & fr2/xyz & fr3/office
& 0000 & 0059 & 0106 & 0169 & 0181 & 0207 \\
\hline

GO-SLAM \cite{Zhang2023goslam}               & \tick & \tick  & \tick  & 0.003 & 0.003 & 0.004 & 0.004 & 0.005 & 0.003 & 0.003 & 0.003 & 0.015 & 0.006 & 0.013 & 0.053 & 0.075 & 0.070 & 0.077 & 0.068 & 0.048 \\
CO-SLAM \cite{Wang2023coslam}               & \xmark & \tick  & \xmark & 0.006 & 0.011 & 0.014 & 0.005 & 0.005 & 0.005 & 0.014 & 0.008 & 0.024 & 0.017 & 0.024 & 0.072 & 0.123 & 0.096 & 0.067 & 0.134 & 0.071 \\
\midrule
\midrule


iMAP* \cite{sucar2021imap,Zhu2022NICE-SLAM}     & \xmark & \xmark  & \xmark & 0.031 & 0.025 & 0.023 & 0.017 & 0.010 & 0.040 & 0.041 & 0.019 & 0.049 & 0.020 & 0.058 & 0.559 & 0.321 & 0.175 & 0.705 & 0.321 & 0.119 \\
NICE-SLAM \cite{Zhu2022NICE-SLAM} & \xmark & \xmark  & \xmark & 0.017 & 0.020 & 0.016 & 0.010 & 0.009 &  0.014 & 0.040 & 0.031 & 0.027    & 0.018   & 0.030      & 0.086 & 0.123 &  0.081 & 0.103 & 0.129 & 0.056 \\     
ESLAM \cite{Johari2023eslam}               & \xmark & \xmark  & \xmark   & 0.007 & 0.007 & 0.006 & 0.006 & 0.006 &  0.006 & 0.007 & 0.006 & 0.025   & 0.011   & 0.024 & 0.073 & 0.085 & 0.075 & 0.065 & 0.090 & 0.057\\
Point-SLAM \cite{Sandstrom2023pointslam}               & \xmark & \xmark  & \xmark  & 0.006 & 0.004 & 0.004 & 0.004 & 0.005 & 0.005 & 0.007 & 0.006 & 0.043 & 0.013 & 0.035 & 0.102 & 0.078 & 0.086 & 0.222 & 0.148 & 0.095 \\    
{ACE-SLAM (ours)} & \xmark & \xmark & \xmark  

& 0.027 & 0.044 & 0.034 & 0.030 & 0.044 & 0.049 & 0.043 & 0.039 & 0.083   & 0.016   & 0.082      & 0.164 & 0.092 &  0.319  &  0.274 & 0.219 & 0.212\\
{ACE-SLAM (w/ \acemlp{})}
& \xmark & \xmark & \xmark  
& 
0.030 & 0.069 & 0.038 & 0.042 & 0.028 & 0.086 & 0.072 & 0.048 & 0.080 & 0.017 & 0.141 & 0.364 & 0.089 & 0.765 & 0.517 & 0.337 & 0.278\\
{ACE-SLAM (w/ SP feat.)}& \xmark & \xmark & \xmark &
0.044 & 0.271 & X & 0.051 & 0.161 & 1.316 & 0.075 & 0.048 & 0.157 & 0.016 & 0.120 & X & 0.103 & 0.233 & 0.257 & X & 0.460\\
\hline
\end{tabular}
\label{tab:static}
\end{table*}

\begin{table}[t]
\centering
\footnotesize
\caption{\textbf{Efficiency Performance Evaluation} on \textit{Replica} Room0 on a NVIDIA RTX 4090 GPU with default configuration.}
\label{tab:compute}
\begin{tabular}{lccccc}
\toprule
Method & FPS ($\downarrow$) & RT-Factor ($\uparrow$) & Map Size ($\downarrow$) \\
\midrule
iMAP* \cite{sucar2021imap,Zhu2022NICE-SLAM}& 0.15 & 0.5\% & 0.99 MB \\ 
Point-SLAM \cite{Sandstrom2023pointslam}& 0.27 & 0.9\% & 27.23 MB \\ 
NICE-SLAM \cite{Zhu2022NICE-SLAM}       & 0.33 & 1.1\% & 95.86 MB \\ 
ESLAM \cite{Johari2023eslam}            & 7.35 & 24.5\% & 45.46 MB \\ 
ACE-SLAM {\scriptsize (Ours)}              & 29.71 & 99.0\% & 1.11 MB \\
ACE-SLAM {\scriptsize (w/ \acemlp{})}              & 29.71 & 99.1\% & 1.01 MB \\
ACE-SLAM {\scriptsize (w/ SP feat.)}               & 29.70 & 99.0 \% & 0.86 MB  \\
\bottomrule
\end{tabular}
\end{table}

\begin{table*}[t]
\centering
\scriptsize
\setlength{\tabcolsep}{2pt}
\renewcommand{\arraystretch}{1.15}
\caption{\textbf{\ac{ATE} \ac{RMSE} [m] ($\downarrow$) in the TUM-RGBD dynamic datasets}. Results for baselines replicated from \cite{Wang2025fndslam,Xu2024nidslam}. X represents failure, N/A indicates no information has not been reported, and (SE) indicates the use of semantic priors.}

\begin{tabular}{l|c|cccccccc}
\toprule
\textbf{Method} & \textbf{SE} &
  \textbf{fr3/s/xyz} & \textbf{fr3/s/rpy} & \textbf{fr3/s/static} & \textbf{fr3/s/half} 
& \textbf{fr3/w/xyz} & \textbf{fr3/w/rpy} & \textbf{fr3/w/static} & \textbf{fr3/w/half} \\
\midrule

iMAP*\cite{sucar2021imap,Zhu2022NICE-SLAM}  & \xmark & {0.420} & {0.364} & {0.074} & {0.812} & {0.410} & {0.834} & 0.102 & 0.638\\ 
NICE-SLAM\cite{Zhu2022NICE-SLAM}            & \xmark & {0.394} & \third{0.099} & {0.029} & {0.569} & {0.302} & {0.724} & 0.092 & {0.629}\\ 
CO-SLAM\cite{Wang2023coslam}                & \xmark & {0.059} & N/A & {0.012} & X & {0.340} & X & {2.518} & N/A \\ 
ESLAM\cite{Johari2023eslam}                 & \xmark & 0.041 & N/A & \third{0.009} & \second{0.033} & {0.432} & {1.044} & {0.294} & N/A \\ 
\midrule
NID-SLAM\cite{Xu2024nidslam}                & \tick & {0.075} & \best{0.086} & {0.019} & {0.105} & \second{0.071} & {0.648} & {0.062} & \best{0.071} \\ 
FND-SLAM\cite{Wang2025fndslam}              & \tick & \best{0.016} & N/A & {0.010} & \best{0.015} & \best{0.013} & \best{0.030} & \best{0.010} & N/A \\ 
\midrule
ACE-SLAM (ours)                                        & \xmark & 
\second{0.037} & 0.119 & \best{0.007} & \third{0.049} & \third{0.072} & \second{ 0.253} & \second{0.012} & \third{0.150} \\ 
ACE-SLAM (w/ \acemlp{})                                        & \xmark & 
\third{0.039} & \second{0.069} & \second{0.008} & \third{0.049} & 0.073 & \third{0.480} & \third{0.017} & \second{0.093}\\
ACE-SLAM (w/ SP feat.)                                        & \xmark & 
0.063 & 0.581 & 0.018 & 0.669 & 0.215 & 1.121 & 0.042 & 0.855\\
\bottomrule
\end{tabular}
\label{tab:dynamic}
\vspace{-0.3cm}
\end{table*}

\begin{figure}
    \centering
    \includegraphics[width=\linewidth]{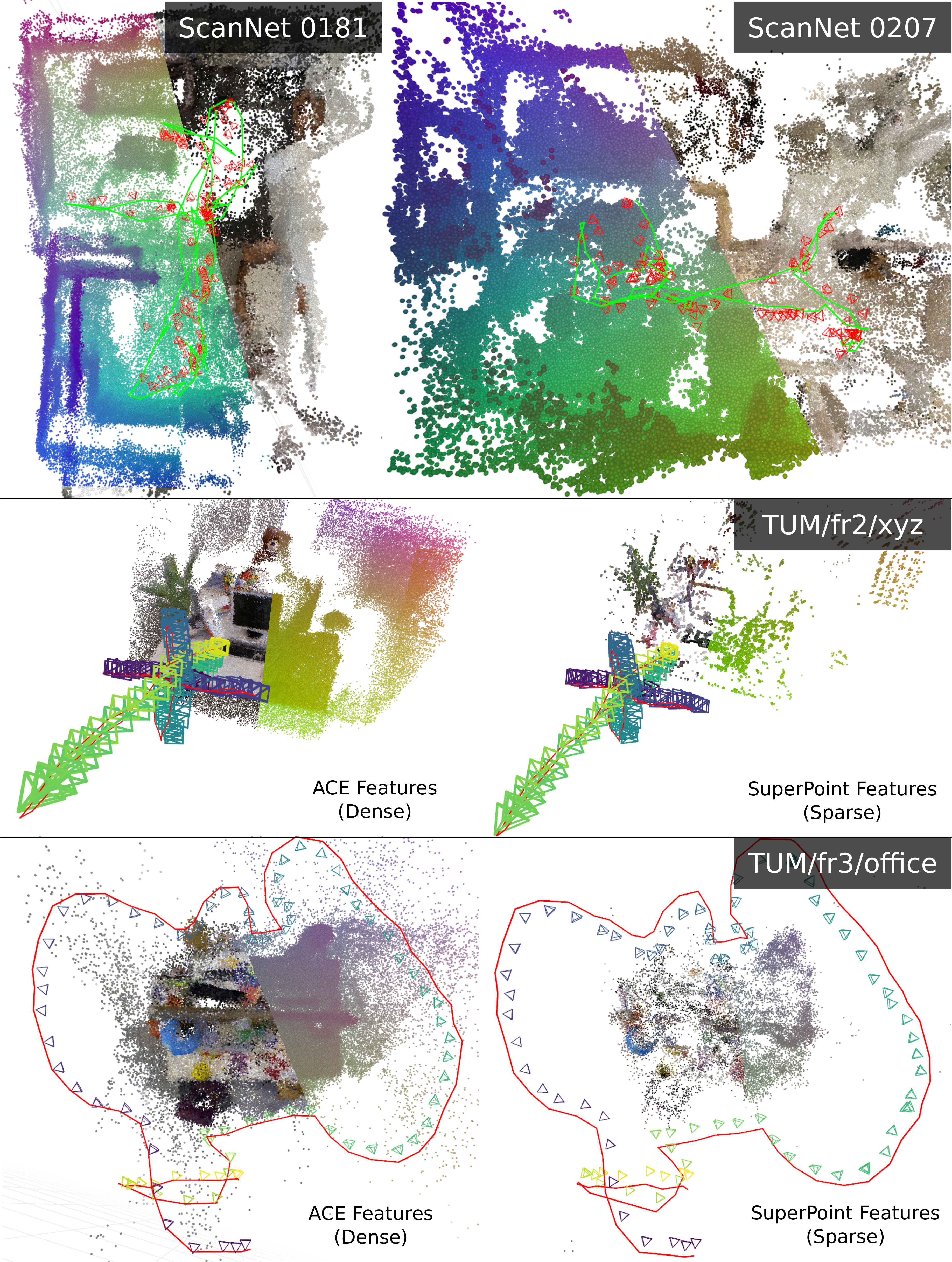}
\caption{\textbf{Qualitative results on static scenes} from \textit{ScanNet} and \textit{TUM-RGBD}.  
Estimated trajectories are shown as frustums, and ground truth trajectories as lines.  
Colored map geometry is overlaid with the corresponding estimated 3D global coordinates.}
    \label{fig:static_qualitative}
\end{figure}

\subsection{Neural Implicit SLAM in Static Scenes}
We evaluate our approach against established  neural implicit RGB-D \ac{SLAM} baselines, including iMAP~\cite{sucar2021imap} (with iMAP* denoting its public implementation from~\cite{Zhu2022NICE-SLAM}), NICE-SLAM~\cite{Zhu2022NICE-SLAM}, ESLAM~\cite{Johari2023eslam}, and Point-SLAM~\cite{Sandstrom2023pointslam}, as summarized in \cref{tab:static}.
We additionally include high-performing pipelines that extend the basic neural implicit RGB-D \ac{SLAM} formulation with specialized modules for improved robustness and efficiency, such as CO-SLAM~\cite{Wang2023coslam} and GO-SLAM~\cite{Zhang2023goslam}.
These systems incorporate components like decoupled tracking front-ends (FE), explicit loop closure (LC), or local/global bundle adjustment (BA). They are shown here for reference only, as their added subsystems make it difficult to isolate the contribution of these auxiliary components from the underlying implicit representation itself, many of which could be also integrated into our framework in future work.

The proposed ACE-SLAM in its main configuration combines dense \ac{ACE} features~\cite{brachmann2023ace,brachmann2024acezero} with the proposed \trimlp{} representation.
It achieves competitive trajectory accuracy (see \cref{fig:static_qualitative}) compared to iMAP* in \textit{Replica} and \textit{TUM} scenes and even surpassing it on parts of \textit{ScanNet}, and approaching NICE-SLAM on several sequences.
Although still below the accuracy of the most recent approaches and pipelines with specialized submodules, our method remains the only one, to the best of our knowledge, capable of strict real-time operation directly from a RGB-D feed, while maintaining an exceptionally compact map representation (see \cref{sec:compute}).

To ablate the effect of our novel \ac{SCR} scene representation, we compare our triplane-based \trimlp{} architecture with the direct regression model from~\cite{brachmann2023ace,brachmann2024acezero}, denoted \acemlp{}, which uses homogeneous parametrization and skip connections.
Both architectures use the same small number of hidden layers. Across all sequences, \trimlp{} vastly outperforms \acemlp{} in trajectory accuracy, specially in large and detailed scenes, highlighting the benefit of structured inductive biases in \ac{SCR}-based mapping for real-time \ac{SLAM}. 
We note that this analysis is limited to real-time SLAM applications, and the applicability of \trimlp{} for other \ac{SCR} tasks, such as large-scale relocalization, remains to be explored in future work.

We further evaluate the flexibility of our framework by comparing dense ACE features~\cite{brachmann2021dsacstar,brachmann2023ace,brachmann2024acezero} with sparse SuperPoint features~\cite{detone2018superpoint}. 
Both perform comparably in small-scale scenes, while ACE features are generally more reliable and accurate overall. 
We hypothesize that dense features offer broader spatial coverage, even with weaker descriptiveness, which benefits mapping within the \ac{SCR} backbone. 
In contrast, sparse SuperPoint features, though highly discriminative for frame-to-frame matching, are more challenging to implicitly match due to lower dimensionality and fewer keypoints. 
These results aims to highlight the flexibility of our framework, which can seamlessly operate on any type of features, either sparse or dense. 

\subsection{Neural Implicit SLAM in Dynamic Scenes}
We evaluate ACE-SLAM on dynamic sequences from \textit{TUM-RGBD}, comparing it against standard baselines as well as dynamic-specific pipelines NID-SLAM~\cite{Xu2024nidslam} and FND-SLAM~\cite{Wang2025fndslam}, which leverage semantic priors (SE) to handle moving objects in the scene.

As shown in \cref{tab:dynamic}, methods that excel in static scenes often fail under even minor dynamics. In contrast, ACE-SLAM, using dense ACE features~\cite{brachmann2024acezero} and the \trimlp{} \ac{SCR} map, performs comparably to or better than pipelines with specialized components, without relying on semantic priors or additional modules. The performance using \acemlp{} follows a similar trend. When employing sparse SuperPoint features, however, accuracy degrades noticeably—likely due to the increased ambiguity of implicit matching under sparsity, as discussed earlier.

\begin{figure}
    \centering
    \includegraphics[width=0.9\linewidth]{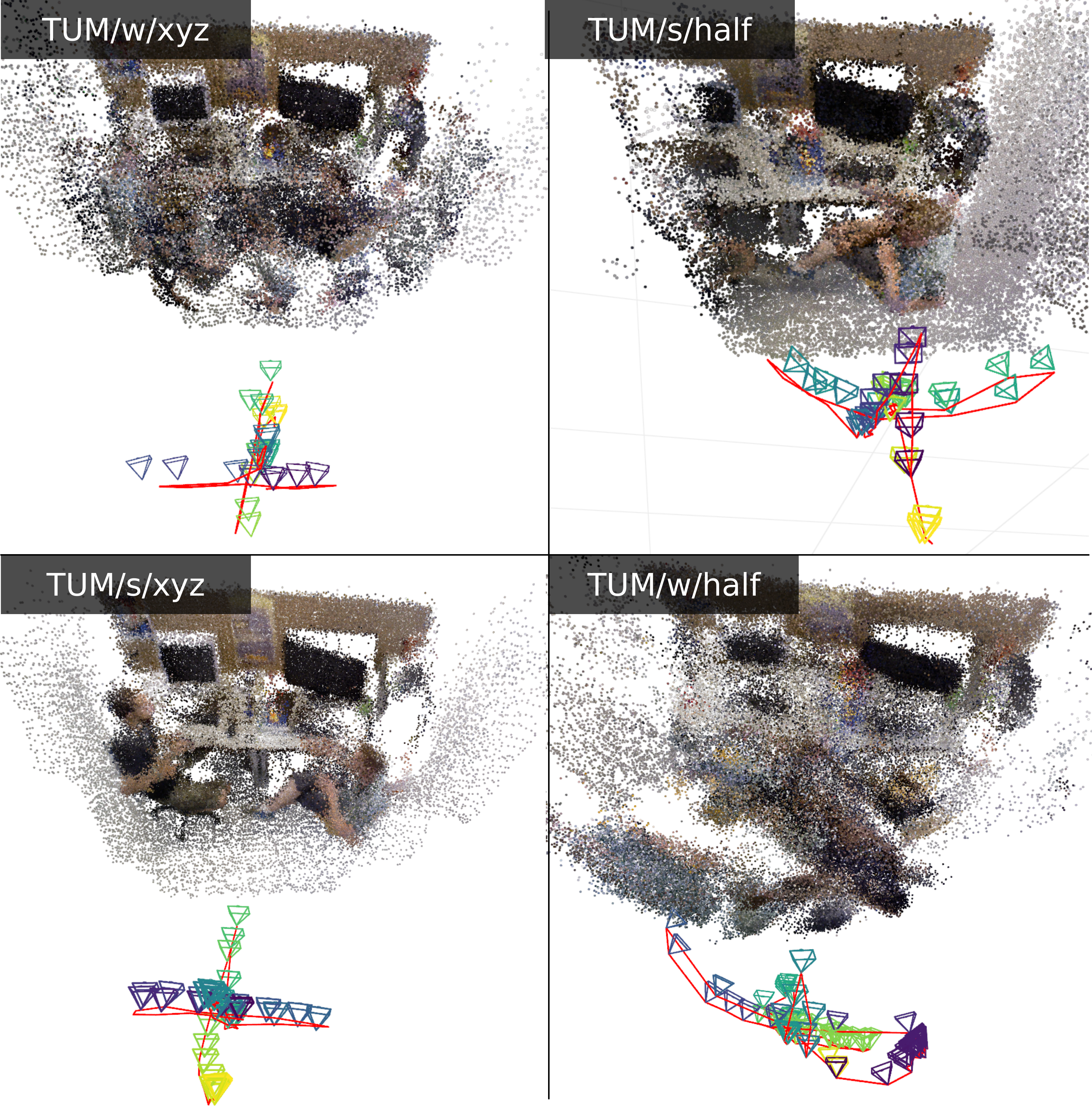}
\caption{\textbf{Qualitative results on dynamic scenes} from \textit{TUM-RGBD}.  
Despite strong scene dynamics such as people moving in front of the camera, the estimated camera trajectory (shown as frustums) closely follows the ground truth (line). Geometric artifacts appear in regions affected by motion, yet ACE-SLAM remains robust and maintains accurate tracking.}
    \label{fig:dynamic_qualitative}
\vspace{-0.6cm}
\end{figure}

This robustness is inherently present in our approach due to the always-on relocalization strategy, which naturally ignores inconsistent or dynamic regions. Importantly, this is achieved while maintaining strict real-time performance, without the added complexity and computational overhead of incorporating specialized submodules for dynamic handling, such as semantic segmentation or optical-flow-based masking. Qualitative examples of this behaviour are shown in \cref{fig:dynamic_qualitative}, demonstrating robustness to moving elements.

\subsection{Computational Performance Analysis}
\label{sec:compute}

While iMAP~\cite{sucar2021imap} targeted near real-time performance, later neural implicit SLAM pipelines often require orders of magnitude more processing time than the original data capture. 
Many report real-time capability based on tracking iteration efficiency, which can be misleading: fast tracking iteration might not necessarily converge faster, and the estimated map must sufficiently well estimated to successfully tracking against it. 
For these reasons, some pipelines decouple tracking and mapping~\cite{Zhang2023goslam,Kong2023vMAP} at the cost of adding complexity to their systems. 
Most pipelines process every frame in a sequence sequentially, taking as long as needed, which does not reflect true online operation (cf. \cite{Tosi2024ReshapingSLAM}). \Eg, iMAP reportedly takes over 10 minutes to map \textit{Replica} Room0~\cite{Kong2023vMAP}, a 1-minute sequence at 30 FPS.
Many methods in the implicit neural SLAM literature cannot tolerate skipping even a single frame (cf. \cite{Wang2023coslam} and \cite{Deng2024PLGSLAM} supp. material), let alone the potentially necessary frame dropping to keep real-time operation \cite{Zhang2023goslam}.

To correctly assess efficiency, we report the so-called equivalent FPS of each method, defined as the total processing time for a sequence divided by its number of frames, and convert this into a real-time factor (RT-Factor) relative to the sequence’s FPS. An RT-Factor of 100\% indicates that processing matches the input frame rate, whereas 50\% means the method requires twice as much time to process the data as it took to generate it.
Our pipeline is explicitly designed for strict real-time operation, as reflected in these metrics: if processing a frame exceeds its capture interval, subsequent frames are skipped to maintain temporal alignment, faithfully mimicking an online camera feed.
In contrast, other methods process frames sequentially without any real-time constraints, which we outperform by up to two orders of magnitude. 
Metrics are reported on the \textit{Replica} Office0 sequence, following standard practice.
We achieve real-time performance while representing the scene with a highly compressed implicit \ac{SCR} map, making tracking and mapping efficient. Localizing a new frame end-to-end takes 11ms (93 FPS equivalent) or 13ms (76 FPS equivalent) using ACE~\cite{brachmann2024acezero} or SuperPoint~\cite{detone2018superpoint} features, respectively. Note that during normal operation, the cost of running the feature extraction is negligible as it is fully parallelized and overlaps with other operations in the pipeline.

These results provide strong evidence supporting \ac{SCR} as a competitive core implicit map representation for neural implicit RGB-D \ac{SLAM}, in contrast to the volumetric, rendering-based approaches that dominate the field. 
We believe this work redefines the state-of-the-art by emphasizing real-time performance, demonstrating that highly efficient \ac{SCR}-based methods can serve as a foundation for future research aimed at improving accuracy without sacrificing online operation.
\section{Conclusions}
We present ACE-SLAM, the first neural implicit RGB-D system that uses \ac{SCR} as its core map representation.
We introduce a novel \ac{SCR} network architecture leveraging explicit 3D coordinate voting, incorporating efficient inductive biases that demonstrably improve performance in real-time \ac{SLAM}.

The proposed system, while conceptually simple, natively supports key \ac{SLAM} functionalities such as implicit loop closure and relocalization without requiring additional complex modules. It is flexible, supporting both sparse and dense feature representations, and operates reliably even in scenes with dynamic elements.
ACE-SLAM compresses room-scale scenes into maps on the order of megabytes, which are privacy-preserving by design (\ie geometric information about the scene can only be extracted if an image of the scene is provided).

Importantly, our system is the first tightly-coupled implicit neural \ac{SLAM} approach to achieve strict real-time performance without relying on additional components, such as decoupled mapping and tracking subsystems, setting new efficiency standards in the state of the art.

{
    \small
    \bibliographystyle{ieeenat_fullname}
    \bibliography{main}

@String(CVPR= {IEEE Conf. Comput. Vis. Pattern Recog.})

@String(ICCV= {Int. Conf. Comput. Vis.})

@String(ECCV= {Eur. Conf. Comput. Vis.})

@String(ICME = {Int. Conf. Multimedia and Expo})

@String(CVPRW= {IEEE Conf. Comput. Vis. Pattern Recog. Worksh.})

@String(threeDV = {Int. Conf. 3D Vision})

@String(CVPR  = {CVPR})

@String(ICCV  = {ICCV})

@String(ECCV  = {ECCV})

@String(ICME  =	{ICME})

@String(CVPRW= {CVPRW})

@inproceedings{sucar2021imap,
  title     = {{iMAP}: Implicit Mapping and Positioning in Real-Time},
  author    = {Edgar Sucar and Shikun Liu and Joseph Ortiz and Andrew J. Davison},
  booktitle = ICCV,
  year      = {2021},
  url       = {http://arxiv.org/abs/2103.12352v2}
}

@inproceedings{Zhu2022NICE-SLAM,
  author    = {Zihan Zhu and Songyou Peng and Viktor Larsson and Weiwei Xu and Hujun Bao and Zhaopeng Cui and Martin R. Oswald and Marc Pollefeys},
  title     = {NICE-SLAM: Neural Implicit Scalable Encoding for SLAM},
  booktitle = CVPR,
  month     = {June},
  year      = {2022},
  doi       = {10.1109/CVPR52688.2022.01245}
}

@inproceedings{Zhu2024NICER,
  author    = {Zihan Zhu and Songyou Peng and Viktor Larsson and Zhaopeng Cui and Martin R. Oswald and Andreas Geiger and Marc Pollefeys},
  title     = {{NICER-SLAM}: Neural Implicit Scene Encoding for RGB SLAM},
  booktitle = threeDV,
  year      = {2024},
  month     = {March},
  pages     = {42--52},
  doi       = {10.1109/3DV62453.2024.00096}
}

@article{Held2025Triangle,
  author    = {Jan Held and Renaud Vandeghen and Adrien Deliege and Abdullah Hamdi and Silvio Giancola and Anthony Cioppa and Andrea Vedaldi and Bernard Ghanem and Andrea Tagliasacchi and Marc Van Droogenbroeck},
  title     = {Triangle Splatting for Real-Time Radiance Field Rendering},
  journal   = {arXiv},
  year      = {2025},
  url       = {https://arxiv.org/abs/2505.19175}
}

@inproceedings{Matsuki2024GaussianSplattingSLAM,
  author    = {Hidenobu Matsuki and Riku Murai and Paul H. J. Kelly and Andrew J. Davison},
  title     = {{Gaussian Splatting SLAM} (MonoGS)},
  booktitle = CVPR,
  month     = {June},
  year      = {2024},
  pages     = {18039--18048},
  doi       = {10.1109/CVPR48898.2024.01836}
}

@inproceedings{Murai2025MASt3RSLAM,
  author    = {Riku Murai and Eric Dexheimer and Andrew J. Davison},
  title     = {{MASt3R-SLAM}: Real-Time Dense SLAM with 3D Reconstruction Priors},
  booktitle = {Proceedings of the IEEE/CVF Conference on Computer Vision and Pattern Recognition (CVPR)},
  year      = {2025},
  month     = {June},
  pages     = {1234--1243},
  doi       = {10.1109/CVPR52688.2025.01234},
  url       = {https://openaccess.thecvf.com/content/CVPR2025/html/Murai_MASt3R-SLAM_Real-Time_Dense_SLAM_with_3D_Reconstruction_Priors_CVPR_2025_paper.html}
}

@inproceedings{Wang2025VGGT,
  author    = {Jianyuan Wang and Minghao Chen and Nikita Karaev and Andrea Vedaldi and Christian Rupprecht and David Novotny},
  title     = {{VGGT}: Visual Geometry Grounded Transformer},
  booktitle = {Proceedings of the IEEE/CVF Conference on Computer Vision and Pattern Recognition (CVPR)},
  year      = {2025},
  doi       = {10.1109/CVPR52688.2025.01234},
  url       = {https://arxiv.org/abs/2503.11651},
  note      = {CVPR 2025 Best Paper Award}
}

@article{Wang2025Pi3,
  author    = {Yifan Wang and Jianjun Zhou and Haoyi Zhu and Wenzheng Chang and Yang Zhou and Zizun Li and Junyi Chen and Jiangmiao Pang and Chunhua Shen and Tong He},
  title     = {$\pi$³: Permutation-Equivariant Visual Geometry Learning},
  journal   = {arXiv preprint arXiv:2507.13347},
  year      = {2025},
  url       = {https://arxiv.org/abs/2507.13347},
  note      = {Accessed: 2025-10-26}
}

@inproceedings{oleynikova2017voxblox,
  author={Oleynikova, Helen and Taylor, Zachary and Fehr, Marius and Siegwart, Roland and  Nieto, Juan},
  booktitle={IEEE/RSJ International Conference on Intelligent Robots and Systems (IROS)},
  title={Voxblox: Incremental 3D Euclidean Signed Distance Fields for On-Board MAV Planning},
  year={2017}
}

@article{Grinvald2019VoxbloxPlusPlus,
  author    = {Margarita Grinvald and Fadri Furrer and Tonci Novkovic and Jen Jen Chung and Cesar Cadena and Roland Siegwart and Juan Nieto},
  title     = {Volumetric Instance-Aware Semantic Mapping and 3D Object Discovery},
  journal   = {IEEE Robotics and Automation Letters},
  year      = {2019},
  volume    = {4},
  number    = {3},
  pages     = {3037--3044},
  doi       = {10.1109/LRA.2019.2923960},
  url       = {https://arxiv.org/abs/1903.0268}
}

@article{MurArtal2017ORB_SLAM2,
  author    = {Raul Mur-Artal and Juan D. Tardós},
  title     = {ORB-SLAM2: an Open-Source SLAM System for Monocular, Stereo, and RGB-D Cameras},
  journal   = {IEEE Transactions on Robotics},
  year      = {2017},
  volume    = {33},
  number    = {5},
  pages     = {1255--1262},
  doi       = {10.1109/TRO.2017.2709694},
  url       = {https://arxiv.org/abs/1610.06475}
}

@article{Qin2018VINSMono,
  author    = {Tong Qin and Peiliang Li and Shaojie Shen},
  title     = {VINS-Mono: A Robust and Versatile Monocular Visual-Inertial State Estimator},
  journal   = {IEEE Transactions on Robotics},
  year      = {2018},
  volume    = {34},
  number    = {4},
  pages     = {1004--1020},
  doi       = {10.1109/TRO.2018.2853729},
  url       = {https://ieeexplore.ieee.org/document/8321902}
}

@article{Campos2021ORB_SLAM3,
  author    = {Carlos Campos and Richard Elvira and Juan J. Gómez Rodríguez and José M. M. Montiel and Juan D. Tardós},
  title     = {ORB-SLAM3: An Accurate Open-Source Library for Visual, Visual-Inertial and Multi-Map SLAM},
  journal   = {IEEE Transactions on Robotics},
  year      = {2021},
  volume    = {37},
  number    = {6},
  pages     = {1874--1890},
  doi       = {10.1109/TRO.2021.3060187},
  url       = {https://arxiv.org/abs/2007.11898}
}

@article{Kerbl2023GaussianSplatting,
  author    = {Bernhard Kerbl and Georgios Kopanas and Thomas Leimk{\"u}hler and George Drettakis},
  title     = {3D Gaussian Splatting for Real-Time Radiance Field Rendering},
  journal   = {ACM Transactions on Graphics},
  year      = {2023},
  volume    = {42},
  number    = {4},
  pages     = {1--15},
  doi       = {10.1145/3592433},
  url       = {https://arxiv.org/abs/2308.04079}
}

@article{Engel2018DSO,
  author    = {Jakob Engel and Vladlen Koltun and Daniel Cremers},
  title     = {Direct Sparse Odometry},
  journal   = {IEEE Transactions on Pattern Analysis and Machine Intelligence},
  year      = {2018},
  volume    = {40},
  number    = {3},
  pages     = {611--625},
  doi       = {10.1109/TPAMI.2017.2762314},
  url       = {https://arxiv.org/abs/1607.02565}
}

@inproceedings{Bloesch2018CodeSLAM,
  author    = {Michael Bloesch and Jan Czarnowski and Ronald Clark and Stefan Leutenegger and Andrew J. Davison},
  title     = {CodeSLAM: Learning a Compact, Optimisable Representation for Dense Visual SLAM},
  booktitle = {Proceedings of the IEEE Conference on Computer Vision and Pattern Recognition (CVPR)},
  year      = {2018},
  pages     = {2560--2568},
  doi       = {10.1109/CVPR.2018.00271},
  url       = {https://arxiv.org/abs/1804.00874}
}

@article{Czarnowski2020DeepFactors,
  author    = {Jan Czarnowski and Tristan Laidlow and Ronald Clark and Andrew J. Davison},
  title     = {DeepFactors: Real-Time Probabilistic Dense Monocular SLAM},
  journal   = {IEEE Robotics and Automation Letters},
  year      = {2020},
  volume    = {5},
  number    = {2},
  pages     = {721--728},
  doi       = {10.1109/LRA.2020.2966791},
  url       = {https://arxiv.org/abs/2001.05049}
}

@inproceedings{Teed2021DROIDSLAM,
  author    = {Zachary Teed and Jia Deng},
  title     = {DROID-SLAM: Deep Visual SLAM for Monocular, Stereo, and RGB-D Cameras},
  booktitle = {Advances in Neural Information Processing Systems (NeurIPS)},
  year      = {2021},
  url       = {https://arxiv.org/abs/2108.10869}
}

@inproceedings{Tateno2017CNN-SLAM,
  author    = {Keisuke Tateno and Federico Tombari and Iro Laina and Nassir Navab},
  title     = {CNN-SLAM: Real-Time Dense Monocular SLAM with Learned Depth Prediction},
  booktitle = {Proceedings of the IEEE Conference on Computer Vision and Pattern Recognition (CVPR)},
  year      = {2017},
  pages     = {6568--6577},
  doi       = {10.1109/CVPR.2017.698},
  url       = {https://arxiv.org/abs/1704.03489}
}

@inproceedings{Wang2024glace,
      author    = {Fangjinhua Wang and Xudong Jiang and Silvano Galliani and Christoph Vogel and Marc Pollefeys},
      title     = {GLACE: Global Local Accelerated Coordinate Encoding},
      booktitle = {Proceedings of the IEEE/CVF Conference on Computer Vision and Pattern Recognition (CVPR)},
      month     = {June},
      year      = {2024}
  }

@inproceedings{brachmann2023ace,
    title={Accelerated Coordinate Encoding: Learning to Relocalize in Minutes using RGB and Poses},
    author={Brachmann, Eric and Cavallari, Tommaso and Prisacariu, Victor Adrian},
    booktitle={CVPR},
    year={2023},
}

@article{brachmann2021dsacstar,
  title={Visual Camera Re-Localization from {RGB} and {RGB-D} Images Using {DSAC}},
  author={Brachmann, Eric and Rother, Carsten},
  journal={TPAMI},
  year={2021}
}

@misc{liu2025mace,
  author = {Mingkai Liu and Dikai Fan and Haohua Que and Haojia Gao and Xiao Liu and Shuxue Peng and Meixia Lin and Shengyu Gu and Ruicong Ye and Wanli Qiu and Handong Yao and Ruopeng Zhang and Xianliang Huang},
  title = {MACE: Mixture-of-Experts Accelerated Coordinate Encoding for Large-Scale Scene Localization and Rendering},
  year = {2025},
  eprint = {2510.14251},
  archivePrefix = {arXiv},
  primaryClass = {cs.CV},
  url = {https://arxiv.org/abs/2510.14251}
}

@inproceedings{jiang2025rscore,
  author    = {Xudong Jiang and Fangjinhua Wang and Silvano Galliani and Christoph Vogel and Marc Pollefeys},
  title     = {R-SCoRe: Revisiting Scene Coordinate Regression for Robust Large-Scale Visual Localization},
  booktitle = {Proceedings of the IEEE/CVF Conference on Computer Vision and Pattern Recognition (CVPR)},
  year      = {2025},
  pages     = {11536--11546},
  doi       = {10.1109/CVPR52734.2025.01077},
  url       = {https://openaccess.thecvf.com/content/CVPR2025/papers/Jiang_R-SCoRe_Revisiting_Scene_Coordinate_Regression_for_Robust_Large-Scale_Visual_Localization_CVPR_2025_paper.pdf}
}

@inproceedings{shotton2013scr_forest,
  author    = {J. Shotton and A. Blake and C. Rother and M. J. Cook and E. Finzi and R. Cipolla},
  title     = {Scene Coordinate Regression Forests for Camera Pose Estimation},
  booktitle = {IEEE Conference on Computer Vision and Pattern Recognition (CVPR)},
  year      = {2013},
  pages     = {2930--2937},
  doi       = {10.1109/CVPR.2013.377},
  url       = {https://openaccess.thecvf.com/content_cvpr_2013/papers/Shotton_Scene_Coordinate_Regression_2013_CVPR_paper.pdf}
}

@Article{Chai2015OptimizedACE,
AUTHOR = {Chai, Xinbo and Yang, Zhen and Tan, Xinrong and Zhu, Mengyang and Zhong, Changbin and Shi, Jianping},
TITLE = {Enhanced Camera Relocalization Through Optimized Accelerated Coordinate Encoding Network and Pose Solver},
JOURNAL = {Sensors},
VOLUME = {25},
YEAR = {2025},
NUMBER = {6},
ARTICLE-NUMBER = {1920},
URL = {https://www.mdpi.com/1424-8220/25/6/1920},
PubMedID = {40293022},
ISSN = {1424-8220},

DOI = {10.3390/s25061920}
}

@inproceedings{Johari2023eslam,
  author    = {Mohammad Mahdi Johari and Camilla Carta and François Fleuret},
  title     = {ESLAM: Efficient Dense SLAM System Based on Hybrid Representation of Signed Distance Fields},
  booktitle = CVPR,
  year      = {2023},
  note      = {Highlight Paper},
  url       = {https://www.idiap.ch/paper/eslam},
}

@inproceedings{Zhang2023goslam,
  author    = {Youmin Zhang and Fabio Tosi and Stefano Mattoccia and Matteo Poggi},
  title     = {GO-SLAM: Global Optimization for Consistent 3D Instant Reconstruction},
  booktitle = ICCV,
  year      = {2023},
  url       = {https://arxiv.org/pdf/2309.02436},
}

@inproceedings{Wang2023coslam,
  author    = {Hengyi Wang and Jingwen Wang and Lourdes Agapito},
  title     = {Co‑SLAM: Joint Coordinate and Sparse Parametric Encodings for Neural Real‑Time SLAM},
  booktitle = CVPR,
  year      = {2023},
  url       = {https://hengyiwang.github.io/projects/CoSLAM.html},
}

@inproceedings{Sandstrom2023pointslam,
  author    = {Erik Sandström and Yue Li and Luc Van Gool and Martin R. Oswald},
  title     = {Point‑SLAM: Dense Neural Point Cloud‑based SLAM},
  booktitle = ICCV,
  year      = {2023},
  url       = {https://arxiv.org/pdf/2304.04278},
}

@article{Deng2024neslam,
  author    = {Tianchen Deng and Yanbo Wang and Hongle Xie and Hesheng Wang and Jingchuan Wang and Danwei Wang and Weidong Chen},
  title     = {NeSLAM: Neural Implicit Mapping and Self‑Supervised Feature Tracking With Depth Completion and Denoising},
  journal   = {arXiv preprint arXiv:2403.20034},
  year      = {2024},
  url       = {https://arxiv.org/pdf/2403.20034}
}

@article{Li2025ddnslam,
  author    = {Mingrui Li and Zhetao Guo and Tianchen Deng and Yiming Zhou and Guangan Jiang and Yangyang Wang and Hongyu Wang},
  title     = {DDN‑SLAM: Real‑time Dense Dynamic Neural Implicit SLAM},
  journal   = {IEEE Robotics and Automation Letters},
  year      = {2025},
  note      = {Preprint available at \url{https://github.com/DrLi‑Ming/DDN‑SLAM}},
}

@article{Jiang2024rodynslam,
  author    = {Haochen Jiang and Yueming Xu and Kejie Li and Jianfeng Feng and Li Zhang},
  title     = {RoDyn‑SLAM: Robust Dynamic Dense RGB‑D SLAM with Neural Radiance Fields},
  journal   = {arXiv preprint arXiv:2407.01303},
  year      = {2024},
  url       = {https://arxiv.org/abs/2407.01303},
}

@inproceedings{Xu2024nidslam,
  author    = {Ziheng Xu and Jianwei Niu and Qingfeng Li and Tao Ren and Chen Chen},
  title     = {NID‑SLAM: Neural Implicit Representation‑based RGB‑D SLAM in Dynamic Environments},
  booktitle = ICME,
  year      = {2024},
  url       = {https://arxiv.org/abs/2401.01189},
}

@ARTICLE{Wang2025fndslam,
  author={Yang, Xianben and Wang, Tao and Wang, Yangyang and Lang, Congyan and Jin, Yi and Li, Yidong},
  journal={IEEE Sensors Journal}, 
  title={FND-SLAM: A SLAM System Using Feature Points and NeRF in Dynamic Environments Based on RGB-D Sensors}, 
  year={2025},
  volume={25},
  number={5},
  pages={8598-8610},
  doi={10.1109/JSEN.2025.3527558}}

@article{GomezOjeda2019PLSLAM,
  author    = {Ruben Gómez-Ojeda and Francisco-Ángel Moreno and David Zuñiga-Noël and Davide Scaramuzza and Javier González-Jiménez},
  title     = {{PL-SLAM}: A Stereo SLAM System Through the Combination of Points and Line Segments},
  journal   = {IEEE Transactions on Robotics},
  year      = {2019},
  volume    = {35},
  number    = {3},
  pages     = {734--746},
  doi       = {10.1109/TRO.2019.2899783},
  url       = {https://doi.org/10.1109/TRO.2019.2899783}
}

@inproceedings{Pumarola2017PLSLAM,
  author    = {Albert Pumarola and Alexander Vakhitov and Antonio Agudo and Alberto Sanfeliu and Francesc Moreno-Noguer},
  title     = {{PL-SLAM}: Real-Time Monocular Visual SLAM with Points and Lines},
  booktitle = {IEEE Int. Conf. on Robotics and Automation (ICRA)},
  year      = {2017},
  pages     = {in press}
}

@inproceedings{Mildenhall2020NeRF,
  author    = {Ben Mildenhall and Pratul P. Srinivasan and Matthew Tancik and Jonathan T. Barron and Ravi Ramamoorthi and Ren Ng},
  title     = {NeRF: Representing Scenes as Neural Radiance Fields for View Synthesis},
  booktitle = {European Conference on Computer Vision (ECCV)},
  year      = {2020},
  pages     = {405–421},
  doi       = {10.1007/978-3-030-58539-6\_24},
  url       = {https://arxiv.org/abs/2003.08934}
}

@article{Muller2022InstantNGP,
  author    = {Thomas Müller and Alex Evans and Christoph Schied and Alexander Keller},
  title     = {Instant Neural Graphics Primitives with a Multiresolution Hash Encoding},
  journal   = {ACM Transactions on Graphics (SIGGRAPH)},
  year      = {2022},
  volume    = {41},
  number    = {4},
  pages     = {Art. 102:1–15},
  doi       = {10.1145/3528223.3530127},
  url       = {https://arxiv.org/abs/2201.05989}
}

@inproceedings{brachmann2024acezero,
  title     = {Scene Coordinate Reconstruction: Posing of Image Collections via Incremental Learning of a Relocalizer},
  author    = {Eric Brachmann and Jamie Wynn and Shuai Chen and Tommaso Cavallari and Áron Monszpart and Daniyar Turmukhambetov and Victor Adrian Prisacariu},
  booktitle = {European Conference on Computer Vision (ECCV)},
  year      = {2024},
  doi       = {10.1007/978-3-031-72992-8\_24},
  url       = {https://arxiv.org/abs/2404.14351}
}

@inproceedings{Lipson2024DPVSLAM,
  author    = {Lahav Lipson and Zachary Teed and Jia Deng},
  title     = {Deep Patch Visual SLAM},
  booktitle = {European Conference on Computer Vision (ECCV)},
  year      = {2024},
  doi       = {10.1007/978-3-031-72627-9\_24},
  url       = {https://arxiv.org/abs/2408.01654}
}

@article{Tosi2024ReshapingSLAM,
  author    = {Fabio Tosi and Youmin Zhang and Ziren Gong and Erik Sandström and Stefano Mattoccia and Martin R. Oswald and Matteo Poggi},
  title     = {How NeRFs and 3D Gaussian Splatting are Reshaping SLAM:a Survey},
  year      = {2024},
  journal   = {arXiv preprint},
  volume    = {abs/2402.13255},
  url       = {https://arxiv.org/abs/2402.13255}
}

@inproceedings{Xu2022PointNeRF,
  author    = {Qiangeng Xu and Zexiang Xu and Julien Philip and Sai Bi and Zhixin Shu and Kalyan Sunkavalli and Ulrich Neumann},
  title     = {Point‑NeRF: Point‑based Neural Radiance Fields},
  booktitle = {Proceedings of the IEEE/CVF Conference on Computer Vision and Pattern Recognition (CVPR)},
  year      = {2022},
  pages     = {5438--5448},
  doi       = {10.1109/CVPR52688.2022.00536},
  url       = {https://openaccess.thecvf.com/content/CVPR2022/papers/Xu_Point-NeRF_Point-Based_Neural_Radiance_Fields_CVPR_2022_paper.pdf}
}

@inproceedings{Chen2022TensoRF,
  author    = {Anpei Chen and Zexiang Xu and Andreas Geiger and Jingyi Yu and Hao Su},
  title     = {TensoRF: Tensorial Radiance Fields},
  booktitle = {European Conference on Computer Vision (ECCV)},
  year      = {2022},
  volume    = {13692},
  pages     = {333--350},
  doi       = {10.1007/978-3-031-19824-3\_20},
  url       = {https://arxiv.org/abs/2203.09517}
}

@article{Wang2024StructerfSLAM,
  author    = {Haocheng Wang and Yanlong Cao and Xiaoyao Wei and Yejun Shou and Lingfeng Shen and Zhijie Xu and Kai Ren},
  title     = {Structerf‑SLAM: Neural Implicit Representation SLAM for Structural Environments},
  journal   = {Computers \& Graphics},
  year      = {2024},
  volume    = {119},
  pages     = {103893},
  doi       = {10.1016/j.cag.2024.103893},
  url       = {https://doi.org/10.1016/j.cag.2024.103893}
}

@inproceedings{Yin2024IBDSLAM,
  author    = {Minghao Yin and Shangzhe Wu and Kai Han},
  title     = {IBD‑SLAM: Learning Image‑Based Depth Fusion for Generalizable SLAM},
  booktitle = {Proceedings of the IEEE/CVF Conference on Computer Vision and Pattern Recognition (CVPR)},
  year      = {2024},
  pages     = {10563--10573},
  doi       = {10.1109/CVPR52688.2024.01018},
  url       = {https://openaccess.thecvf.com/content/CVPR2024/papers/Yin_IBD-SLAM_Learning_Image-Based_Depth_Fusion_for_Generalizable_SLAM_CVPR_2024_paper.pdf}
}

@inproceedings{Deng2024PLGSLAM,
  author    = {Tianchen Deng and Guole Shen and Tong Qin and Jianyu Wang and Wentao Zhao and Jingchuan Wang and Danwei Wang and Weidong Chen},
  title     = {PLGSLAM: Progressive Neural Scene Representation with Local to Global Bundle Adjustment},
  booktitle = {Proceedings of the IEEE/CVF Conference on Computer Vision and Pattern Recognition (CVPR)},
  year      = {2024},
  pages     = {19657--19666},
  doi       = {10.1109/CVPR52734.2024.019657},
  url       = {https://openaccess.thecvf.com/content/CVPR2024/papers/Deng_PLGSLAM_Progressive_Neural_Scene_Represenation_with_Local_to_Global_Bundle_CVPR_2024_paper.pdf}
}

@inproceedings{Wang2025CUT3R,
  author    = {Qianqian Wang and Yifei Zhang and Aleksander Holynski and Alexei A. Efros and Angjoo Kanazawa},
  title     = {CUT3R: Continuous 3D Perception Model with Persistent State},
  booktitle = {Proceedings of the IEEE/CVF Conference on Computer Vision and Pattern Recognition (CVPR)},
  year      = {2025},
  note      = {Oral presentation},
  url       = {https://cut3r.github.io/}
}

@inproceedings{Liu2025SLAM3R,
  author    = {Yuzheng Liu and Siyan Dong and Shuzhe Wang and Yingda Yin and Yanchao Yang and Qingnan Fan and Baoquan Chen},
  title     = {SLAM3R: Real‑Time Dense Scene Reconstruction from Monocular RGB Videos},
  booktitle = {Proceedings of the IEEE/CVF Conference on Computer Vision and Pattern Recognition (CVPR)},
  year      = {2025},
  pages     = {16651--16662},
  doi       = {10.1109/CVPR52734.2025.01552},
  url       = {https://openaccess.thecvf.com/content/CVPR2025/papers/Liu_SLAM3R_Real-Time_Dense_Scene_Reconstruction_from_Monocular_RGB_Videos_CVPR_2025_paper.pdf}
}

@article{Fischler1981RANSAC,
  author  = {M. A. Fischler and R. C. Bolles},
  title   = {Random sample consensus: A paradigm for model fitting with applications to image analysis and automated cartography},
  journal = {Communications of the ACM},
  volume  = {24},
  number  = {6},
  pages   = {381--395},
  month   = {Jun},
  year    = {1981},
  doi     = {10.1145/358669.358692}
}

@inproceedings{Brachmann2019NeuralGuidedRANSAC,
  author    = {Eric Brachmann and Carsten Rother},
  title     = {Neural‑Guided RANSAC: Learning Where to Sample Model Hypotheses},
  booktitle = {Proceedings of the IEEE/CVF International Conference on Computer Vision (ICCV)},
  year      = {2019},
  pages     = {4321--4330},
  url       = {https://openaccess.thecvf.com/content_ICCV_2019/papers/Brachmann_Neural-Guided_RANSAC_Learning_Where_to_Sample_Model_Hypotheses_ICCV_2019_paper.pdf}
}

@inproceedings{Brachmann2019ExpertSampleConsensus,
  author    = {Eric Brachmann and Carsten Rother},
  title     = {Expert Sample Consensus Applied to Camera Re‑Localization},
  booktitle = {Proceedings of the IEEE/CVF International Conference on Computer Vision (ICCV)},
  year      = {2019},
  pages     = {7524--7533},
  url       = {https://openaccess.thecvf.com/content_ICCV_2019/papers/Brachmann_Expert_Sample_Consensus_Applied_to_Camera_Re-Localization_ICCV_2019_paper.pdf}
}

@article{Wang2024HSCNetPlusPlus,
  author    = {Shuzhe Wang and Zakaria Laskar and Iaroslav Melekhov and Xiaotian Li and Yi Zhao and Giorgos Tolias and Juho Kannala},
  title     = {HSCNet++: Hierarchical Scene Coordinate Classification and Regression for Visual Localization with Transformer},
  journal   = {International Journal of Computer Vision},
  year      = {2024},
  volume    = {132},
  pages     = {2530--2550},
  doi       = {10.1007/s11263-023-01982-9},
  url       = {https://link.springer.com/article/10.1007/s11263-023-01982-9}
}

@inproceedings{Bruns2025ACEG,
  author    = {Leonard Bruns and Axel Barroso‑Laguna and Tommaso Cavallari and Áron Monszpart and Sowmya Munukutla and Victor Adrian Prisacariu and Eric Brachmann},
  title     = {{ACE‑G}: Improving Generalization of Scene Coordinate Regression Through Query Pre‑Training},
  booktitle = {Proceedings of the IEEE/CVF International Conference on Computer Vision (ICCV)},
  year      = {2025},
  pages     = {26751--26761},
  url       = {https://openaccess.thecvf.com/content/ICCV2025/papers/Bruns_ACE-G_Improving_Generalization_of_Scene_Coordinate_Regression_Through_Query_Pre-Training_ICCV_2025_paper.pdf}
}

@inproceedings{Bian2025SceneCoordinateReconstructionPriors,
  author    = {Wenjing Bian and Axel Barroso‑Laguna and Tommaso Cavallari and Victor Adrian Prisacariu and Eric Brachmann},
  title     = {Scene Coordinate Reconstruction Priors},
  booktitle = {Proceedings of the IEEE/CVF International Conference on Computer Vision (ICCV)},
  year      = {2025},
  url       = {https://openaccess.thecvf.com/content/ICCV2025/papers/Bian_Scene_Coordinate_Reconstruction_Priors_ICCV_2025_paper.pdf}
}

@inproceedings{Klein2007PTAM,
  author    = {Georg Klein and David Murray},
  title     = {Parallel Tracking and Mapping for Small AR Workspaces},
  booktitle = {Proceedings of the International Symposium on Mixed and Augmented Reality (ISMAR)},
  year      = {2007},
  pages     = {225--234},
  doi       = {10.1109/ISMAR.2007.4538852},
  url       = {https://doi.org/10.1109/ISMAR.2007.4538852}
}

@article{Umeyama1991LeastSquares,
  author    = {Shinji Umeyama},
  title     = {Least-Squares Estimation of Transformation Parameters Between Two Point Patterns},
  journal   = {IEEE Transactions on Pattern Analysis and Machine Intelligence},
  year      = {1991},
  volume    = {13},
  number    = {4},
  pages     = {376--380},
  doi       = {10.1109/34.88573},
  url       = {https://doi.org/10.1109/34.88573}
}

@article{Straub2019Replica,
  author    = {Julian Straub and Thomas Whelan and Lingni Ma and Yufan Chen and Erik Wijmans and Simon Green and Jakob J. Engel and Raul Mur‑Artal and Carl Ren and Shobhit Verma and Anton Clarkson and Mingfei Yan and Brian Budge and Yajie Yan and Xiaqing Pan and June Yon and Yuyang Zou and Kimberly Leon and Nigel Carter and Jesus Briales and Tyler Gillingham and Elias Müeggler and Luis Pesqueira and Manolis Savva and Dhruv Batra and Hauke M. Strasdat and Renzo De Nardi and Michael Goesele and Steven Lovegrove and Richard Newcombe},
  title     = {The {\emph{Replica}} Dataset: A Digital Replica of Indoor Spaces},
  journal   = {arXiv preprint arXiv:1906.05797},
  year      = {2019},
  url       = {https://arxiv.org/abs/1906.05797}
}

@inproceedings{Dai2017ScanNet,
  author    = {Angela Dai and Angel X. Chang and Manolis Savva and Maciej Halber and Thomas Funkhouser and Matthias Nießner},
  title     = {ScanNet: Richly‑Annotated 3D Reconstructions of Indoor Scenes},
  booktitle = {Proceedings of the IEEE/CVF Conference on Computer Vision and Pattern Recognition (CVPR)},
  year      = {2017},
  pages     = {665--674},
  url       = {https://openaccess.thecvf.com/content_cvpr_2017/papers/Dai_ScanNet_Richly-Annotated_3D_CVPR_2017_paper.pdf}
}

@inproceedings{Sturm2012BenchmarkRGBD,
  author    = {J\"{u}rgen Sturm and Nico Engelhard and Friedrich Endres and Wolfram Burgard and Daniel Cremers},
  title     = {A Benchmark for the Evaluation of RGB‑D SLAM Systems},
  booktitle = {Proceedings of the IEEE/RSJ International Conference on Intelligent Robots and Systems (IROS)},
  year      = {2012},
  pages     = {573–580},
  url       = {https://vision.in.tum.de/data/datasets/rgbd-dataset/}
}

@misc{grupp2017evo,
  title={evo: Python package for the evaluation of odometry and SLAM.},
  author={Grupp, Michael},
  howpublished={\url{https://github.com/MichaelGrupp/evo}},
  year={2017}
}

@inproceedings{Kong2023vMAP,
  author    = {Xin Kong and Shikun Liu and Marwan Taher and Andrew J. Davison},
  title     = {vMAP: Vectorised Object Mapping for Neural Field SLAM},
  booktitle = {Proceedings of the IEEE/CVF Conference on Computer Vision and Pattern Recognition (CVPR)},
  year      = {2023},
  pages     = {952--961},
  url       = {https://openaccess.thecvf.com/content/CVPR2023/papers/Kong_vMAP_Vectorised_Object_Mapping_for_Neural_Field_SLAM_CVPR_2023_paper.pdf}
}

@inproceedings{DeTone2018SuperPoint,
  author    = {Daniel DeTone and Tomasz Malisiewicz and Andrew Rabinovich},
  title     = {SuperPoint: Self-Supervised Interest Point Detection and Description},
  booktitle = {Proceedings of the IEEE/CVF Conference on Computer Vision and Pattern Recognition Workshops (CVPRW)},
  year      = {2018},
  pages     = {224--236},
  url       = {https://arxiv.org/abs/1712.07629}
}
}


\end{document}